\documentclass[journal,twoside,web]{ieeecolor}
\usepackage{generic}
\usepackage{cite}
\usepackage{amsmath,amssymb,amsfonts}
\usepackage{graphicx}
\usepackage{textcomp}
\usepackage{hyperref}

\hypersetup{
    colorlinks=true,
    linkcolor=blue,
    filecolor=magenta,      
    urlcolor=blue,
}

\newcommand*{\change}{\textcolor{black}}

\newcommand*{\changesec}{\textcolor{black}}

\usepackage{graphics}           
\usepackage{times}              
\usepackage{amsmath}            
\usepackage{amssymb}            
\usepackage{graphicx}
\usepackage{algorithm}
\usepackage[noend]{algpseudocode}
\usepackage{booktabs}
\usepackage{color}
\definecolor{instructioncolor}{rgb}{.5,.5,.5}

\usepackage[font=small]{caption}

\def\secref#1{Sec.~\ref{#1}}
\def\figref#1{Fig.~\ref{#1}}
\def\tabref#1{Tab.~\ref{#1}}
\def\eqref#1{Eq.~(\ref{#1})}

\makeatletter
\usepackage{xspace}
\DeclareRobustCommand\onedot{\futurelet\@let@token\@onedot}
\def\@onedot{\ifx\@let@token.\else.\null\fi\xspace}

\def\etal{{et al}\onedot}
\makeatother

\usepackage{array}
\newcolumntype{L}[1]{>{\raggedright\let\newline\\\arraybackslash\hspace{0pt}}m{#1}}
\newcolumntype{C}[1]{>{\centering\let\newline\\\arraybackslash\hspace{0pt}}m{#1}}
\newcolumntype{R}[1]{>{\raggedleft\let\newline\\\arraybackslash\hspace{0pt}}m{#1}}

\newcommand{\RR}{\mathbb{R}}

\renewcommand{\b}[1]{\mbox{\boldmath$#1$}}

\newcommand{\set}[1]{\mathcal{#1}} 	

\renewcommand{\vec}[1]{{\b #1}}

\usepackage{makecell}
\usepackage{subfigure}
\usepackage{multirow}
\usepackage{rotating}  
\makeatletter
\def\maketag@@@#1{\hbox{\m@th\normalfont\normalsize#1}}
\makeatother

\def\BibTeX{{\rm B\kern-.05em{\sc i\kern-.025em b}\kern-.08em
		T\kern-.1667em\lower.7ex\hbox{E}\kern-.125emX}}
\title{\LARGE \bf CVTNet: A Cross-View Transformer Network for LiDAR-Based Place Recognition in Autonomous Driving Environments}

\author{Junyi Ma, Guangming Xiong, Jingyi Xu, Xieyuanli Chen* 
  \thanks{This work was supported by the National Natural Science Foundation of China under Grant 52372404.}
  \thanks{
  J. Ma, G. Xiong, and J. Xu are with the Beijing Institute of Technology. X. Chen is with the National University of Defense Technology.}
  \thanks{$^*$Corresponding author email: xieyuanli.chen@nudt.edu.cn}
}

\begin{document}
\maketitle


\begin{abstract}
LiDAR-based place recognition (LPR) is one of the most crucial components of autonomous vehicles to identify previously visited places in GPS-denied environments. Most existing LPR methods use mundane representations of the input point cloud without considering different views, which may not fully exploit the information from LiDAR sensors. In this article, we propose a \underline{c}ross-\underline{v}iew \underline{t}ransformer-based network, dubbed CVTNet, to fuse the range image views (RIVs) and bird's eye views (BEVs) generated from the LiDAR data. It extracts correlations within the views using intra-transformers and between the two different views using inter-transformers. Based on that, our proposed CVTNet generates a yaw-angle-invariant global descriptor for each laser scan end-to-end online and retrieves previously seen places by descriptor matching between the current query scan and the pre-built database. We evaluate our approach on three datasets collected with different sensor setups and environmental conditions. The experimental results show that our method outperforms the state-of-the-art LPR methods with strong robustness to viewpoint changes and long-time spans. \changesec{Furthermore, our approach has better real-time performance that can run faster than the typical LiDAR frame rate does.} 
The implementation of our method is released as open source at: \url{https://github.com/BIT-MJY/CVTNet}.

\end{abstract}


\section{Introduction}
\label{sec:intro}

Place recognition provides the current global location of the vehicle in the previously seen environments, which is an important component of robotic simultaneous localization and mapping (SLAM) and global localization. During online operation, it retrieves the reference scan in the database most similar to the current query by directly regressing the similarity~\cite{chen2021auro, ma2022tii} or descriptor matching~\cite{uy2018pointnetvlad, liu2019lpd, vid2022icra}. LiDAR-based place recognition methods~\cite{ma2022ral, kim2018scan, cattaneo2022tro} can be applied to large-scale outdoor environments due to their robustness to illumination and weather changes. Different representation forms of LiDAR data have been exploited as input for LPR methods, such as 3D point clouds~\cite{uy2018pointnetvlad, vid2022icra}, voxels~\cite{fan2022svt, Komorowski2021wacv, chang2020spoxelnet}, normal distributions transform cells~\cite{zaganidis2019semantically, zhou2021ndt}, bird's eye views (BEVs)~\cite{kim2018scan, wang2020intensity, luo2021ral}, and range image views (RIVs)~\cite{chen2021auro, ma2022ral}. 
However, few methods fuse them together to exploit the advances of different views. \change{
Yin \etal~\cite{yin2021fusionvlad} proposed FusionVLAD combining the different input forms, but they simply used fully connected layers to fuse features, thus not yet thoroughly digging out the inner relations between different representations.}

\begin{figure}
\vspace{0.2cm}
  \centering
  \includegraphics[width=1\linewidth]{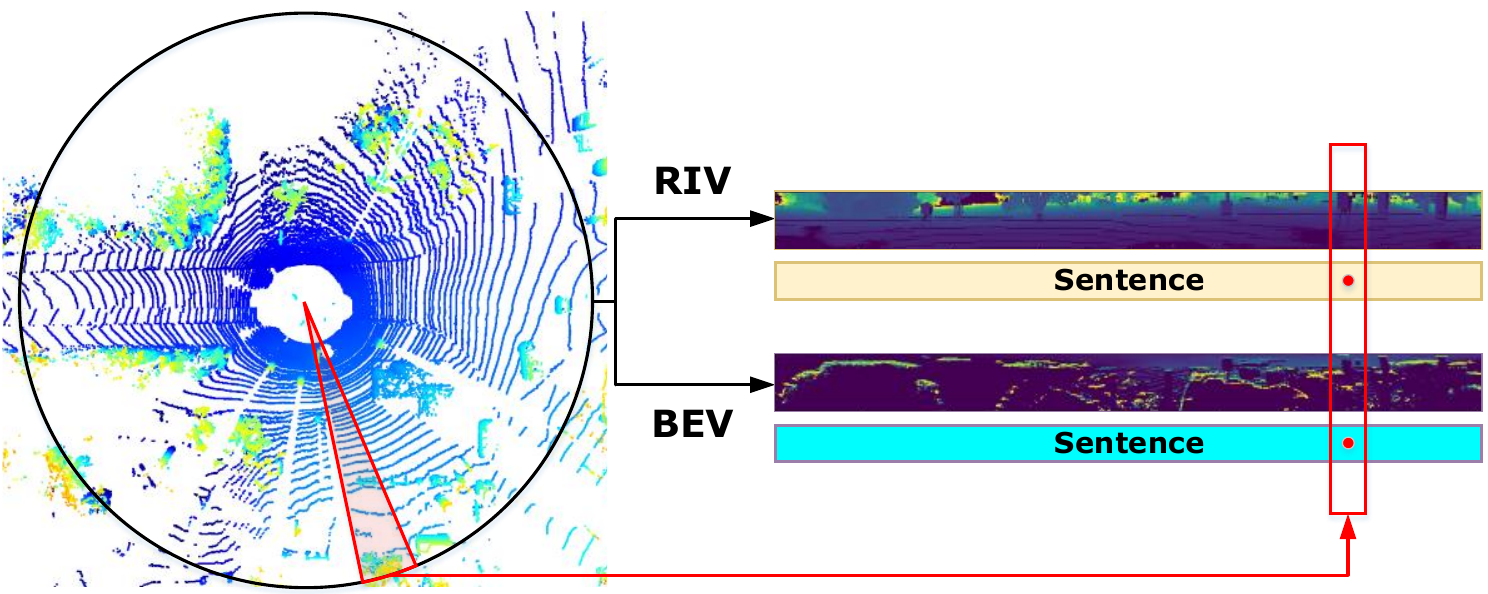}
  \caption{The correspondence between the RIV and BEV leads to aligned feature sentences used by our proposed CVTNet.}
  \label{fig:motivation}
  \vspace{-0.6cm}
\end{figure}

To fill this gap, \changesec{a novel cross-view transformer network named CVTNet is proposed for LiDAR-based place recognition in this work.} It utilizes the combination of multi-layer RIVs and BEVs generated from 3D LiDARs mounted on autonomous vehicles. \changesec{Exploiting the correspondence between the RIVs and BEVs of the same LiDAR scan as shown in~\figref{fig:motivation}, these two input views are first compressed into aligned sentence-like features.} Then, our proposed CVTNet uses intra- and inter-transformers to extract the correlation within and cross the aligned sentences from different views. Based on the proposed novel transformer network, our method generates a yaw-angle-invariant global descriptor for each place in an end-to-end manner and, in the end, achieves robust place recognition based on descriptor matching.

The main contribution of this article is an end-to-end transformer network that fuses RIVs and BEVs of LiDAR data to achieve reliable long-term place recognition. Unlike the existing LPR methods using mundane input forms, our method exploits both RIVs and BEVs of LiDAR data. Furthermore, we extend the vanilla single-channel RIVs and BEVs to multi-layer formats, which enables our transformer network to reweigh the LiDAR data from different ranges and heights, thus generating more reliable and distinctive features. To our best knowledge, this is the first work to dig out the latent alignment of RIV and BEV features from LiDAR point clouds and use the property to generate distinguishable representations for places. Our proposed CVTNet is also the first network exploiting cross-attention to fuse the multi-view LiDAR data and generate yaw-angle-invariant global descriptors for LiDAR place recognition. Benefiting from the devised yaw-angle-invariant architecture, CVTNet is robust to viewpoint changes and achieves reliable place recognition even when the car drives in opposite directions, \changesec{which is thoroughly evaluated on the NCLT, KITTI, and self-recorded datasets.} The experimental results show that our proposed method outperforms the state-of-the-art methods in terms of both loop closure detection and place recognition.  \changesec{The ablation studies on the intra- and inter-transformer modules and the validation experiments on yaw-angle-invariance are further provided.}

In sum, we make three claims that our approach is able to
(i)~achieve state-of-the-art performance in place recognition and loop closure detection in outdoor large-scale environments using multi-view LiDAR data,
(ii)~recognize places well despite different viewpoints based on the proposed yaw-angle-invariant architecture,
(iii)~achieve online operation with a runtime less than 50\,ms faster than the typical LiDAR frame rate.
All claims are supported by our experimental evaluation on
multiple datasets. 

\changesec{The remainder of this paper is organized as follows. Section II will investigate the related works. In Section III, the architecture of our proposed CVTNet and the related mathematical derivation will be described in detail. Section IV will present the experimental setups and results as well as corresponding analysis. Finally, Section V concludes the paper.}

\section{Related Work}
\label{sec:related}

Place recognition has been widely investigated in robotics and computer vision~\cite{ ye2022tii, arandjelovic2016netvlad}. In this article, we only focus on LiDAR-based place recognition (LPR).

LPR is a classic research topic, and many traditional hand-crafted methods have been proposed~\cite{kim2018scan, li2018tii, wang2020lidar, wang2020intensity, luo2021ral}. Recently, more and more learning-based methods have shown significant improvements with the development of neural network techniques. Parts of existing works use the raw point cloud or its voxelized and segmented forms as the input of the model. For example, Uy \etal~\cite{uy2018pointnetvlad} first combine the commonly used PointNet~\cite{qi2017pointnet} with a differentiable VLAD architecture~\cite{arandjelovic2016netvlad}, providing the basic framework for the following methods to extract global descriptors from point clouds. \change{Liu \etal~\cite{liu2019lpd} proposed LPD-Net which utilizes a graph neural network to aggregate multiple local features extracted from point clouds. SegMap proposed by 
Dub\'{e} \etal~\cite{dube2018ijrr_new} first extracts the segments from the point clouds as the input of the devised convolution neural network.} LoGG3D-Net by Vidanapathirana \etal~\cite{vid2022icra} exploits sparse convolution directly on raw point clouds, and considers both local consistency loss and scene-level loss. In contrast, MinkLoc3D by Komorowski~\cite{Komorowski2021wacv} combines the sparse voxelized representation and convolutions to tackle the unordered set problem. More recently, LCDNet proposed by Cattaneo \etal~\cite{cattaneo2022tro} uses an adapted PV-RCNN architecture to extract features that are used to generate global descriptors for point clouds and estimate the relative poses between the query and reference scan.

Although point cloud-based LPR methods have achieved reasonable results, most of them cannot operate online, thus hindering the deployment in real-world applications. To tackle this, many other works use image-like representations of LiDAR data, i.e., RIVs or BEVs, to improve the recognition efficiency and investigate more significant properties. For example, OverlapNet~\cite{chen2021auro} utilizes the range images to directly regress out the overlap values and relative yaw angles between query and reference scans. Following the overlap conception, OverlapTransformer~\cite{ma2022ral} takes advantage of the naturally yaw-angle-equivariance and builds a yaw-angle-invariant architecture. SphereVLAD++ by Zhao \etal~\cite{zhao2022ral} also uses RIVs to extract viewpoint-invariant global descriptors. Different from RIVs, BEVs lose some fine depth information but maintain stable height information, which is useful for long-term localization. For example, Scan Context~\cite{kim2018scan} and Intensity Scan Context~\cite{wang2020intensity} both utilize the top-down view to encode each point cloud to a feature matrix. LiDAR-Iris \cite{wang2020lidar} further extends the context-based methods to the frequency domain using Fourier transform. Inspired by LiDAR-Iris, BVMatch~\cite{luo2021ral} also refers to the advantages of frequency domain, and extracts keypoints and BVFT descriptors for bag-of-words based retrieval. CCL~\cite{cui2023ccl} proposes to use contrastive learning to extract more distinctive descriptor of places to achieve better LPR results.

All the LPR methods mentioned above only use one single input format of point clouds without combining different formats to further enhance the representative ability of the extracted features and descriptors. FusionVLAD by Yin \etal~\cite{yin2021fusionvlad} is the first work that considers the inputs from different views of raw point clouds. However, it only applies fully connected networks to combine the extracted features, which can not fully exploit different information from multi-views. Transformer~\cite{vaswani2017attention} has recently been deployed to better extract features for place recognition~\cite{zhou2021ndt, ma2022ral, fan2022svt}. However, there is no work successfully using transformers to fuse the features from different input forms or views. In this article, we propose a novel transformer network to extract the correlation within and between RIVs and BEVs. We explicitly align the RIV and BEV features from the same LiDAR data, thus better finding the relations between the sentence-like features from the dual views using cross-transformer and extracting more distinct global descriptors for more accurate place recognition. 
 
\begin{figure*}[!t]
  \centering
  \includegraphics[width=\linewidth]{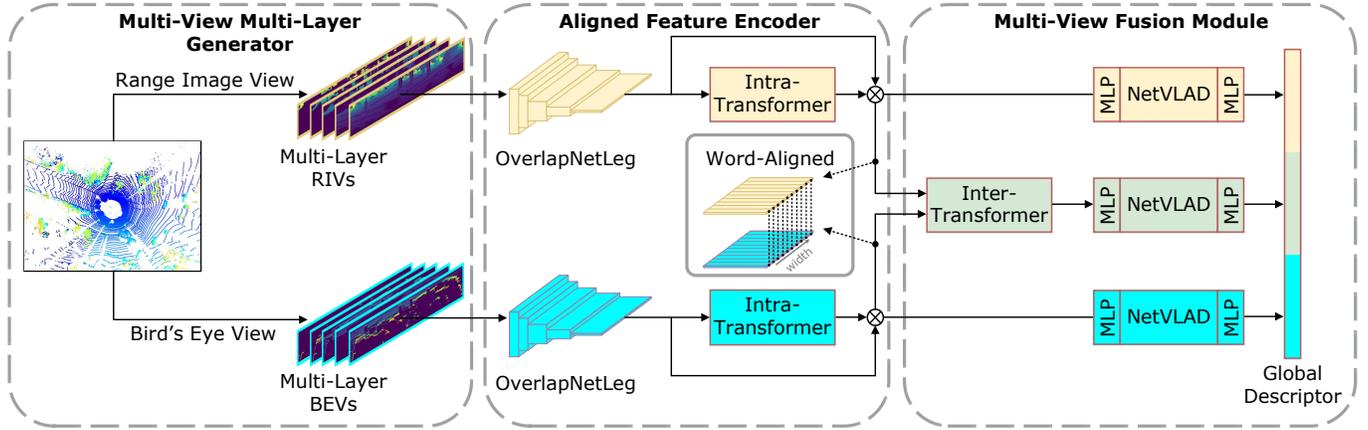}
  \caption{Pipeline overview of our proposed CVTNet. It first generates multi-layer RIV and BEV images for the two separate branches, and encodes the aligned sentence-like features respectively using the combination of OverlapNetLeg and the intra-transformer module. Then, the inter-transformer module is utilized to extract the inner correlation cross views. In the end, NetVLAD with MLP is exploited on three branches respectively to generate the final yaw-angle-invariant global descriptors for fast retrieval of place recognition.}
  \label{fig:pipeline}
  \vspace{-0.4cm}
\end{figure*}

\section{Proposed Method} 
\label{sec:overlap_network}


The overview of our proposed method is depicted in~\figref{fig:pipeline}. CVTNet consists of three modules, a multi-view multi-layer generator (MMG), an aligned feature encoder (AFE), and a multi-view fusion module (MVF). MMG first splits the region by range and height intervals, and conducts spherical and top-down projection respectively to generate multi-layer input data (see~\secref{sec:mmg}). The following AFE takes the multi-layer RIVs and BEVs as inputs and uses a fully convolutional encoder to generate sentence-like features. It then applies an intra-transformer to extract the inner correlation of the compressed features of each view (see~\secref{sec:afe}). Finally, the enhanced feature volumes are fed to the MVF module. It first uses an inter-transformer to fuse the aligned features from different views and then applies NetVLAD with multi-layer perceptrons (MLPs) on both single-view and fused features to generate the final 1-D global descriptor (see~\secref{sec:mvf}). The final descriptor is invariant against yaw-angle rotation (see~\secref{sec:yawai}). We train our network using the triplet loss with overlap labels to better distinguish the positive and negative examples (see~\secref{sec:train}).

\subsection{Multi-View Multi-Layer Generator}
\label{sec:mmg}

It is important for LPR methods to extract discriminative features from different scans. Existing works obtain RIVs or BEVs using single spherical or top-down projection leads to data loss, thus degrading retrieving performance. To fully exploit all information from LiDAR data while maintaining fast speed, we propose to use multiple projections at different ranges/heights and form multi-layer inputs based on space splits. 
By doing that, our proposed network learns to weigh information from different ranges/heights differently, thus extracting more representative features from scans. 

More specifically, we propose the multi-view multi-layer generator.
For RIVs in spherical coordinate systems, it first discretizes the region by preset range intervals $\{s_0, s_1, s_2, s_3, ..., s_q\}$, obtaining the split space $E^r=\{E_1^r, E_2^r, E_3^r, ..., E_q^r\}$. \change{Then, it applies the spherical projection on laser points of the input point cloud $\set{P}$ within different range intervals to generate multi-layer range images.} Similarly, for BEVs in the Euclidean coordinates, it discretizes space by height intervals $\{t_0, t_1, t_2, t_3, ..., t_q\}$ to get $E^b=\{E_1^b, E_2^b, E_3^b, ..., E_q^b\}$ and generates multi-layer BEV images for all the intervals.

The correspondence between a LiDAR point $\vec{p}_{k} \in \set{P}, \vec{p}_k = (x_k, y_k, z_k)$ and the pixel coordinates $(u_k^r,v_k^r)$ in the corresponding RIV $\set{R}_i$ can be represented as:
\begin{align}
  \left( \begin{array}{c} u_k^r \vspace{0.0em}\\ v_k^r \end{array}\right) & = \left(\begin{array}{cc} \frac{1}{2}\left[1-\arctan(y_k, x_k)/ \pi\right] w   \vspace{0.5em} \\
      \left[1 - \left(\arcsin(z_k/r_k) + \mathrm{f}_{\mathrm{up}}\right)/ \mathrm{f}\right] h\end{array} \right), 
  \label{eq:rivprojection}
\end{align}
where $r_k = ||\vec{p}_k||_2 \in [s_{i-1}, s_{i}]$ is the range measurement of the corresponding point,~$\mathrm{f} = \mathrm{f}_{\mathrm{up}} + \mathrm{f}_{\mathrm{down}}$ is the vertical field-of-view of the sensor, and~$w, h$ are the width and height of the resulting range image. 

We project the same LiDAR points to BEVs by the top-down projection. The correspondence between the same LiDAR point $\vec{p}_{k} \in \set{P}$ and the pixel coordinates $(u_k^b,v_k^b)$ in the corresponding BEV image $\set{B}_j$ can be represented as:
\begin{align}
  \left( \begin{array}{c} u_k^b \vspace{0.0em}\\ v_k^b \end{array}\right) & = \left(\begin{array}{cc} \frac{1}{2}\left[1-\arctan(y_k, x_k)/ \pi\right] w   \vspace{0.5em} \\
      \left[r'_k/\mathrm{f}' \right] h\end{array} \right), \label{eq:bevprojection}
\end{align}
where $r'_k = ||(x_k, y_k)||_2 \in [t_{j-1}, t_{j}]$ and $\mathrm{f}'$ is the maximum sensing range. 

\begin{figure}[t]
  \centering
  \includegraphics[width=1\linewidth]{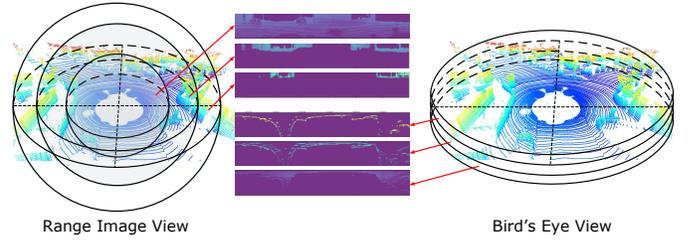}
  \caption{Multi-layer generation for RIVs (left) and BEVs (right).}
  \label{fig:multi-layer}
  \vspace{-0.4cm}
\end{figure}

In this work, we let the RIV and BEV images have the same width and height for better alignment, leading to $u_k^r=u_k^b$. This leads to the fact that the columns of the RIV and BEV images with the same index are both from the same spatial sector, as illustrated in \figref{fig:motivation}. 
Finally, MMG generates the multi-layer RIVs $\RR=\{\set{R}_i\}$ and multi-layer BEVs $\mathbb{B}=\{\set{B}_j\}$, as illustrated in \figref{fig:multi-layer}. 

\subsection{Aligned Feature Encoder}
\label{sec:afe}
To better fuse different representations of LiDAR data, our CVTNet utilizes an aligned feature encoder to extract coarse features of each representation and align them for later fusion. It first concatenates all RIVs $\RR$ and BEVs $\mathbb{B}$ along the channel dimension obtaining $\set{R}$ and $\set{B}$ with the size of $q \times h\times w $. Then it applies two OverlapNetLeg~\cite{chen2021auro,ma2022ral} modules with the same architecture to compress inputs into sentence-like features. The OverlapNetLeg is a full convolutional encoder only to compress RIVs and BEVs in the vertical dimension without changing the width dimension to avoid the discretization error to the yaw equivariance. The output coarse features from each branch are then enhanced by an intra-transformer. We denote the enhanced features as $\set{A}_0^r = \text{AFE}_r(\set{R})$ for the RIV branch, and $\set{A}_0^b = \text{AFE}_b(\set{B})$ for the BEV branch, both with the size of $c \times 1\times w $, where $c$ refers to the feature channel number. 

As proved in~\cite{ma2022ral}, the features from the RIV branch are yaw-angle-equivariant:
\begin{align} 
 \set{A}_0^r C = \text{AFE}_r(\Pi_r(R_\theta \set{P})),
 \label{eq:afer}
\end{align} 
where $\Pi_r(\cdot)$ is the spherical projection detailed in~\eqref{eq:rivprojection}, $C$ represents the column shift of the feature by matrix right multiplication, $\theta$ is the yaw angle change, and $R_\theta$ is the yaw rotation matrix of the point cloud $\set{P}$. 

In this article, we further show that our proposed BEV branch is also yaw-angle-equivariant. According to the top-down projection $\Pi_b(\cdot)$ detailed in~\eqref{eq:bevprojection}, a yaw rotation of a point corresponds to a specific horizontal shift on BEV as:  
\begin{align}
  \left( \begin{array}{c} u_k^{b \prime} \vspace{0.0em}\\ v_k^{b \prime} \end{array}\right) & = \left(\begin{array}{cc} u_k^b + s   \vspace{0.5em} \\
  v_k^b  \end{array} \right), \hspace{0.5em}
  s = \frac{1}{2}\left[1-\theta\pi^{-1}\right] w,
  \label{eq:projection2}
\end{align}
\begin{align}  
  \set{B} C = \Pi_b(R_\theta \set{P}),
\end{align}
where $C$ and $R_\theta$ are the same column shift and yaw rotation. 

As the OverlapNetLeg and intra-transformer have proved to be yaw-angle-equivariant~\cite{ma2022ral}, for the BEV branch, we therefore also obtain the yaw-angle-equivariant features:
\begin{align} 
	\set{A}_0^b C = \text{AFE}_b(\Pi_b(R_\theta \set{P})),
	\label{eq:afeb}
\end{align} 

It is crucial to align the features from different modalities before using the cross-transformer to fuse the different views and extract the cross-modal correlations \cite{Zadeh2018aaai, Pham2019aaai}. To this end, we first make sure the inputs of inter-transformer from the two different views are spatially aligned. We let the input RIV and BEV images have the same width and height, and thus $u_k^r=u_k^b$ holds as clarified in~\secref{sec:mmg}. The same laser point is projected to the same column of the two images, leading to one space sector corresponding to the same column index of the two images. To keep the alignments, we further design our network architecture only to compress the height dimension and maintain the information from one space sector aggregated to the same column of the two output features. We concatenate the output feature of the OverlapNetLeg and the feature enhanced by intra-transformer along the channel dimension. Therefore, the output features $\set{A}_0^r$ and $\set{A}_0^b$ of AFE are devised to be word-aligned along the width dimension. 

\subsection{Multi-View Fusion Module}
\label{sec:mvf}

\begin{figure}[t]
  \centering
  \includegraphics[width=0.8\linewidth]{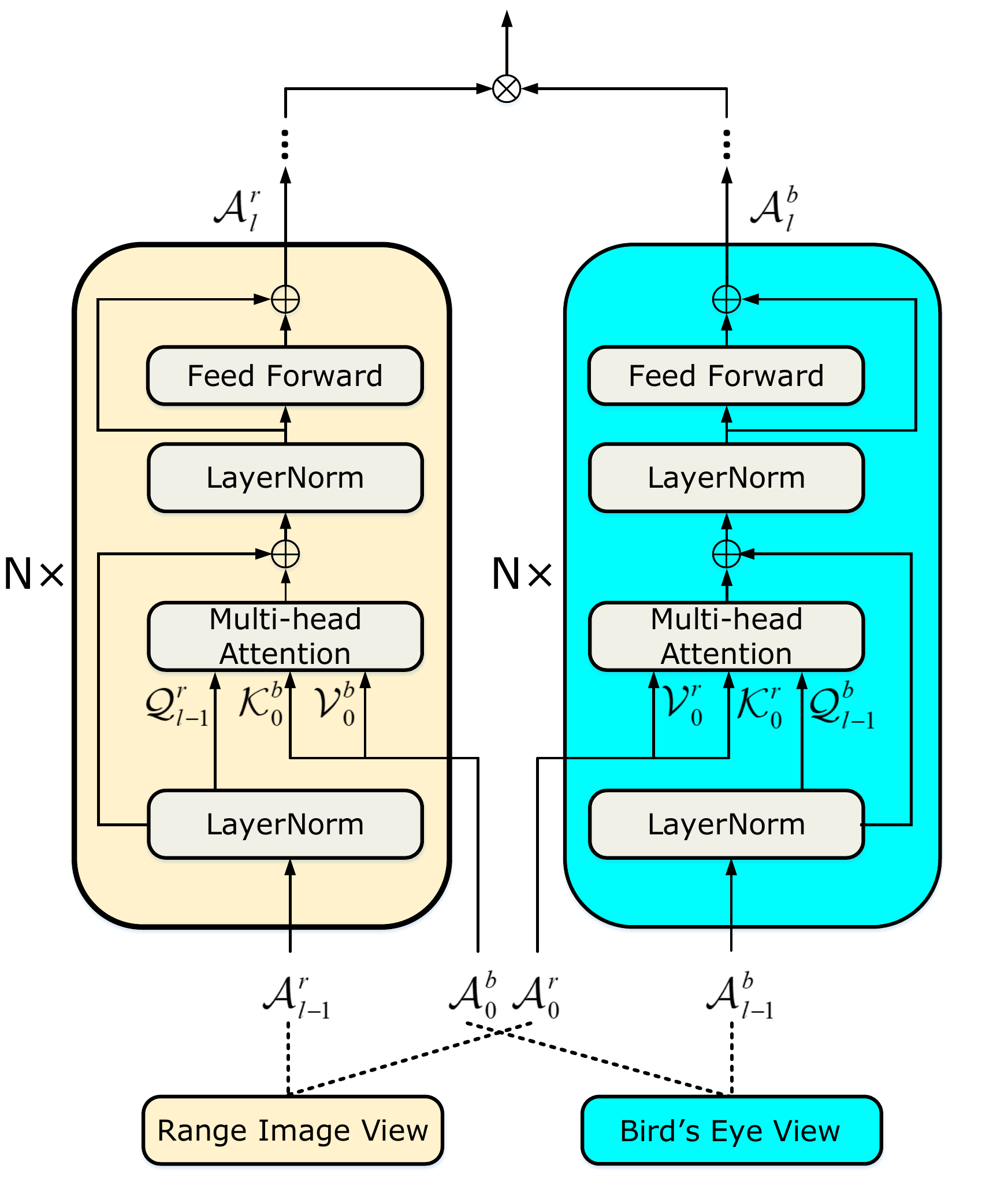}
  \caption{The inter-transformer module.}
  \label{fig:inter}
  \vspace{-0.5cm}
\end{figure}

Our CVTNet fuses the multi-view features in the devised multi-view fusion module. It consists of an inter-transformer and three NetVLAD-MLPs combos~\cite{arandjelovic2016netvlad} as depicted in~\figref{fig:pipeline}.  
It first uses the proposed inter-transformer module to fuse the features from two branches of different views, as detailed in \figref{fig:inter}. In the inter-transformer module, there are also two branches with $l$ stacked transformer blocks to extract the correlation of different views by using the query feature from one view, and the key and value features from the other view. 

For the RIV branch, we denote the feature volume extracted by the multi-head self-attention (MHSA) of the  $l_{th}$ transformer block as $\set{A}_l^r$, and the cross-attention mechanism of the transformer can then be formulated as:
\begin{equation}
\small{
  \bar{\set{A}}_l^r =\text{Attention}(\set{Q}_{l-1}^r, \set{K}_0^b, \set{V}_0^b) 
  =\text{softmax} \Big(\frac{\set{Q}_{l-1}^r {\set{K}_0^b}^{T}}{\sqrt{d_k}}\Big) \set{V}_0^b \vspace{0.5em} ,
 \label{eq:mhsa_r1}
}
\end{equation}
where $\set{Q}_{l-1}^r$ is substantially the query split of the layer-normalized $\set{A}_{l-1}^r$ from the RIV, and $\set{A}_0^r$ is the output feature from the AFE in the RIV branch. $\set{K}_0^b, \set{V}_0^b$ are the key and value splits of the layer-normalized $\set{A}_0^b$, which is the output feature from the AFE in the BEV branch. $d_k$ represents the dimension of splits.
$\bar{\set{A}}_l^r$ is then fed into the feed-forward network (FFN) and layer normalization (LN) to generate the attention-enhanced feature $\set{A}_l^r$ of the $l_{th}$ transformer block, which can be formulated as:
\begin{equation}
\small{
 \set{A}_l^r =\text{FFN}(\text{LN}(\bar{\set{A}}_l^r+\text{LN}(\set{A}_{l-1}^r))) + \text{LN}(\bar{\set{A}}_l^r+\text{LN}(\set{A}_{l-1}^r)).
 \label{eq:mhsa_r2}
}
\end{equation}

For the BEV branch, we conduct the same operations symmetrically:
\begin{equation}
\small{
  \bar{\set{A}}_l^b=\text{Attention}(\set{Q}_{l-1}^b, \set{K}_0^r, \set{V}_0^r) 
  =\text{softmax} \Big(\frac{\set{Q}_{l-1}^b {\set{K}_0^r}^{T}}{\sqrt{d_k}}\Big) \set{V}_0^r ,
}
\end{equation}
\vspace{-0.3cm}
\begin{equation}
\small{
 \set{A}_l^b =\text{FFN}(\text{LN}(\bar{\set{A}}_l^b+\text{LN}(\set{A}_{l-1}^b))) + \text{LN}(\bar{\set{A}}_l^b+\text{LN}(\set{A}_{l-1}^b)).
 \label{eq:mhsa_b1}
}
\end{equation}

We then concatenate $\set{A}_l^r$ with $\set{A}_l^b$ from the last transformer block and obtain the intermediate fused feature $\set{A}_l^f$.

In the end, our MVF uses the NetVLAD-MLPs combos on both the RIV and BEV features $\set{A}_0^r$ and $\set{A}_0^b$, as well as the cross-fused feature $\set{A}_l^f$. 
The NetVLAD-MLPs has been widely used in LPR \cite{zhou2021ndt, cattaneo2022tro, liu2019lpd} to generate the global descriptor.
We also exploit three NetVLAD-MLPs combos to compress the features from two intra-transformers and one cross-view inter-transformer into the global descriptors $[g^r, g^b, g^{f}]$. 

\subsection{Yaw-Angle Invariance}
\label{sec:yawai}

We further mathematically prove that the multi-view fused descriptor generated by CVTNet is yaw-angle-invariant, which makes our method very robust to viewpoint changes. 
We have shown in~\secref{sec:afe} that the outputs of the aligned feature encoder are yaw-angle-equivariant in \eqref{eq:afer} and \eqref{eq:afeb}. As introduced by~\cite{uy2018pointnetvlad}, NetVLAD is permutation-invariant, and the column shift of its input caused by the yaw rotation of the raw point cloud can be regarded as the re-order of a set of 1-D vectors over the channel dimension~\cite{ma2022ral}. Therefore, the outputs of NetVLAD in the RIV and BEV branches are not affected by the yaw rotation, and the global descriptors $g^r$ and $g^b$ are yaw-angle-invariant. In the rest, we prove that the fused cross-view feature $g^{f}$ is also yaw-angle-invariant.

Note that $\set{A}_0^r$ and $\set{A}_0^b$ share the same shift $C$ caused by one yaw rotation of one LiDAR data, which has been introduced in~\secref{sec:afe}. Therefore, for the first transformer block in the inter-transformer after rotation, the \eqref{eq:mhsa_r1} becomes:
\begin{align}
  \widetilde{\set{A}}_1^r&=\text{Attention}(C \set{Q}_{0}^r, C \set{K}_0^b, C \set{V}_0^b),  \nonumber \\
  &=\text{softmax} \Big(\frac{C \set{Q}_{0}^r {(C \set{K}_0^b)}^{T}}{\sqrt{d_k}}\Big) C \set{V}_0^b, \nonumber \\ 
  &= C\, \text{softmax} \Big(\frac{\set{Q}_{0}^r {\set{K}_0^b}^{T}}{\sqrt{d_k}}\Big) \set{V}_0^b =C \bar{\set{A}}_1^r \vspace{0.5em}.
 \label{eq:mhsa_rb1}
\end{align}
Similarly, for the following stacked cross-transformer blocks we also have:
\begin{align}
  \widetilde{\set{A}}_l^r&=\text{Attention}(C \set{Q}_{l-1}^r, C \set{K}_0^b, C \set{V}_0^b), \nonumber \\ 
  &=\text{softmax} \Big(\frac{C \set{Q}_{l-1}^r {(C \set{K}_0^b)}^{T}}{\sqrt{d_k}}\Big) C \set{V}_0^b,  \nonumber \\ 
  &= C\, \text{softmax} \Big(\frac{\set{Q}_{l-1}^r {\set{K}_0^b}^{T}}{\sqrt{d_k}}\Big) \set{V}_0^b =C \bar{\set{A}}_l^r \vspace{0.5em},
 \label{eq:mhsa_rb2}
\end{align}
which means the outputs of the MHSA of the cross-view transformer blocks are yaw-angle-equivariant. Note that in \eqref{eq:mhsa_rb1} and \eqref{eq:mhsa_rb2} the shift $C$ is left-multiplied since the first dimension of the input feature is width dimension and the second one is embedding dimension.
Because FFN and LN are also yaw-angle-equivariant towards features \cite{ma2022ral}, and the last concatenation is operated on the channel dimension, the inter-transformer outputs yaw-angle-equivariant cross-view feature volumes. Then the NetVLAD with MLP is used to generate the yaw-angle-invariant global descriptor $g^{f}$.
Since the three parts are all yaw-angle-invariant, our final global descriptor is yaw-angle-invariant. 

\subsection{Network Training}
\label{sec:train}

Following~\cite{chen2021auro, ma2022ral}, we use more suitable overlap between pairs of scans to supervise the network rather than the ground truth distances. For each training tuple, we utilize one query descriptor $\set{G}_{a}$, $k_p$ positive descriptors $\{\set{G}_p\}$, and $k_n$ negative descriptors $\{\set{G}_n\}$ to compute lazy triplet loss:
\begin{align}
& \set{L}_T(\set{G}_q,\{\set{G}_p\},\{\set{G}_n\})=    \nonumber \\
& k_p(\alpha+\max_{p}(d(\set{G}_q,\set{G}_p))) 
- \sum_{k_n}(d(\set{G}_q,\set{G}_n)),
\label{eq:tripletloss}
\end{align}
where $\alpha$ is the margin to avoid negative loss and $d(\cdot)$ is the squared Euclidean distance. In our work, A pair of scans with overlap larger than 0.3 is regarded as a positive sample, otherwise a negative sample. In addition, we set $k_p=6$, $k_n=6$, and $\alpha=0.5$ for the triplet loss.
We use the triplet loss to make the distance between the query and all sampled negative global descriptors much larger than the distance between the query and the hardest positive global descriptors.



\section{Experimental Evaluation}
\label{sec:exp}
The experimental evaluation is designed to showcase the performance of our approach and to support the claims that our approach is able to:
(i)~achieve state-of-the-art performance in place recognition and loop closure detection in outdoor large-scale environments using multi-view LiDAR data,
(ii)~recognize places well despite different viewpoints based on the proposed yaw-angle-invariant architecture,
(iii)~achieve fast operation with a runtime less than 50\,ms.

\begin{figure}[t]
  \centering
  \includegraphics[width=0.8\linewidth]{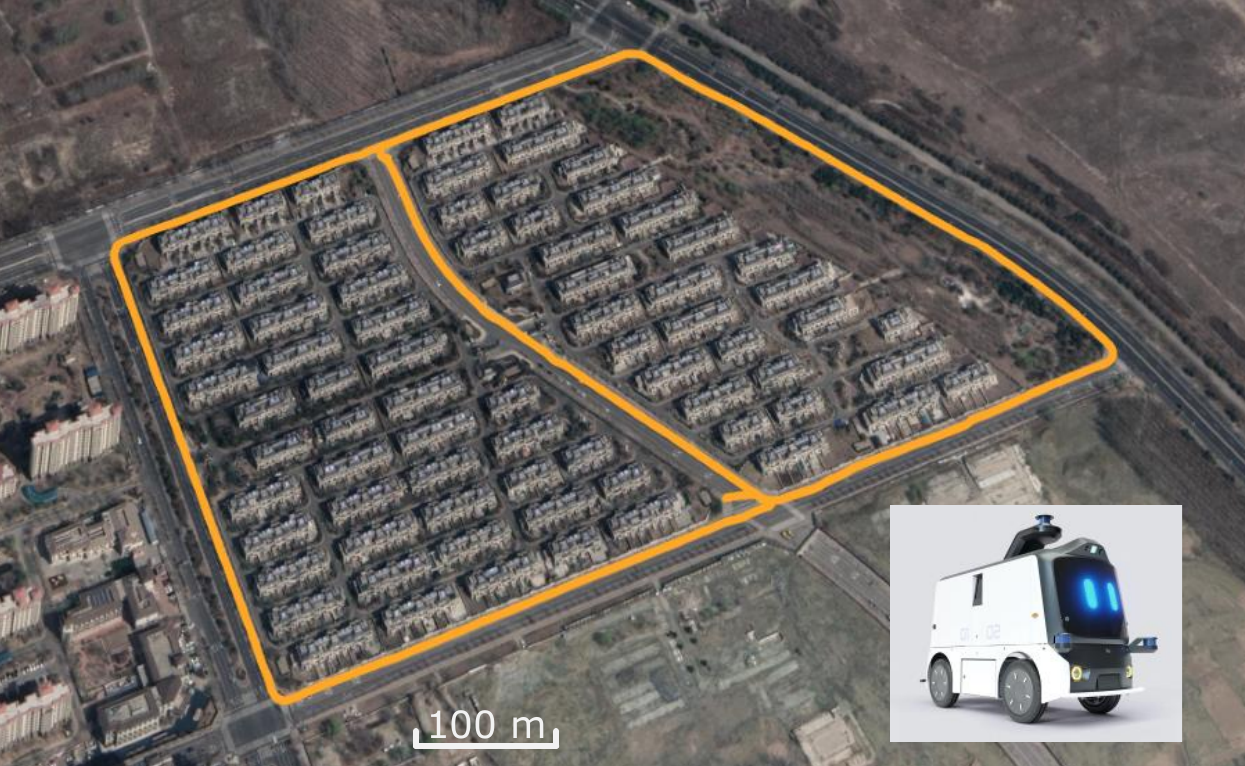}
  \caption{Our self-recorded dataset collected by an autonomous vehicle in the urban environments.}
  \label{fig:haomo_dataset}
  \vspace{-0.4cm}
\end{figure}

\subsection{Implementation and Experimental Setup}
To evaluate localization performance of our proposed method, we utilize three datasets with different environmental conditions and sensor setups, including publicly available NCLT dataset~\cite{carlevaris2016ijrr} and KITTI odometry sequences~\cite{geiger2012cvpr}, as well as our self-recorded datasets in urban driving environments. \change{Our self-recorded dataset is collected by an AGV equipped with a LiDAR sensor (HESAI PandarXT, 32-beam), a wide-angle camera (SENSING SG2, HFOV $106^{\circ}$, VFOV $56^{\circ}$), an RTK GNSS (Asensing INS570D, $0.1^{\circ}$ error in roll/pitch, $0.2^{\circ}$ error in yaw, $2$ cm $+$ $1$ ppm in RTK position), a mini computer (Intel Xeon E, $3.5$ GHz, $80$ W), and a Nivida Tesla T4 (16-GB memory, $70$ W), as shown in \figref{fig:haomo_dataset}.}

\subsubsection{RIV Inputs}
For RIV inputs of NCLT and self-recorded datasets with 32-beam LiDAR data, we set range intervals as $\{[0\,\text{m},15\,\text{m}], [15\,\text{m}, 30\,\text{m}], [30\,\text{m}, 45\,\text{m}], [45\,\text{m}, 60\,\text{m}]\}$ to generate $4$-layer range images. We also generate the range image using all the laser points within $60\,\text{m}$, and concatenate it with $4$-layer range images to get the final RIV input with the size of $5\times32\times 900$. For KITTI sequences, we set the range intervals as $\{[0\,\text{m},15\,\text{m}], [15\,\text{m}, 30\,\text{m}], [30\,\text{m}, 45\,\text{m}], [45\,\text{m}, 80\,\text{m}]\}$ to keep the same maximum range as existing works~\cite{chen2021auro, ma2022ral} for fair comparisons. 
\subsubsection{BEV Inputs}
For the BEV inputs of the NCLT dataset and self-recorded dataset, we set height intervals to $\{[-4\,\text{m},0\,\text{m}], [0\,\text{m}, 4\,\text{m}], [4\,\text{m}, 8\,\text{m}], [8\,\text{m}, 12\,\text{m}]\}$ to generate $4$-layer BEV images, which are also concatenated with the BEV image projected from all the laser points within $60\,\text{m}$. We set the height intervals of KITTI sequences to $\{[-3\,\text{m},-1.5\,\text{m}], [-1.5\,\text{m}, 0\,\text{m}], [0\,\text{m}, 1.5\,\text{m}], [1.5\,\text{m}, 5\,\text{m}]\}$, adapted to its data distribution. The multi-layer BEVs have the same size as the multi-layer RIVs.
\subsubsection{CVTNet Configuration}
\change{Our CVTNet uses the same OverlapNetLeg as OverlapTransformer~\cite{ma2022ral}.
For all the transformer modules, we set the embedding dimension $d_{model}=256$, the number of heads $n_{head}=4$, and the intermediate dimension of the feed-forward layer $d_\mathit{ffn}=1024$. \change{We use one self-attention transformer block in each intra-transformer module following~\cite{ma2022ral}, and two cross-transformer blocks in the inter-transformer module, which is validated as the best configuration in the following ablation study.}
For all the NetVLAD modules, we set the intermediate feature dimension $d_{inter}=1024$, the output feature dimension $d_{output}=256$, and the number of clusters $d_{K}=64$. The output descriptor of CVTNet is a vector of size $1 \times 768$. 
} \changesec{For evaluation conducted on each dataset, our proposed CVTNet is trained for 30 epochs, using the ADAM
optimizer to update the weights of the network with an initial learning rate of 5e-6 and a weight decay of 0.9 applied every
5 steps. Other related training configuration has been clarified in \secref{sec:train}. The test setups follow the previous work \cite{ma2022seqot}, which will also be introduced in the following sections. 
We will release the implementation of CVTNet as open source to provide more network details and evaluation configurations.
}
\subsubsection{Baseline Configuration}
\change{In the following experiments, we compare our proposed CVTNet with both state-of-the-art methods PointNetVLAD~\cite{uy2018pointnetvlad}, Scan Context \cite{kim2018scan} without shift matching for real-time retrieval, MinkLoc3D~\cite{Komorowski2021wacv}, OverlapTransformer~\cite{ma2022ral}, and the other multi-view LPR method FusionVLAD~\cite{yin2021fusionvlad}.
The descriptors generated by all the above-mentioned learning-based baseline methods are with the size of $1 \times 256$ except for FusionVLAD which generates the fused $1 \times 2048$ descriptor, all following their original configurations. The feature matrix generated by Scan Context is set to $20 \times 60$. Note that we reimplement FusionVLAD according to its original paper since it is not open-source, and the other parameter configurations of all the baseline methods in the experiments also follow their original papers.}

\subsection{Evaluation for Place Recognition}
\label{sec:epr}

\begin{table*}[h]
\setlength{\tabcolsep}{3.3pt}
\center
\renewcommand\arraystretch{1.2}
\caption{Comparison of recognition performance on the NCLT dataset}
\begin{tabular}{lcccccccccccc}
\toprule
\multirow{2}{*}{Approach} & \multicolumn{3}{c}{2012-02-05}                      & \multicolumn{3}{c}{2012-06-15}                      & \multicolumn{3}{c}{2013-02-23}                      & \multicolumn{3}{c}{2013-04-05}                      \\ \cline{2-13} 
                          & AR@1        & AR@5        & AR@20       & AR@1        & AR@5        & AR@20       & AR@1        & AR@5        & AR@20       & AR@1        & AR@5        & AR@20       \\ \hline
Scan Context~\cite{kim2018scan}              & 0.767          & 0.836          & 0.909          & 0.581          & 0.637          & 0.724          & 0.481          & 0.564          & 0.726          & 0.418          & 0.496          & 0.649          \\ \hline
PointNetVLAD~\cite{uy2018pointnetvlad}              & 0.746          & 0.823          & 0.875          & 0.612          & 0.720          & 0.782          & 0.469          & 0.604          & 0.719          & 0.449          & 0.576          & 0.683          \\ \hline
MinkLoc3D~\cite{Komorowski2021wacv}        & 0.802          & 0.864          & 0.926          & 0.630          & 0.685          & 0.774          & 0.507          & 0.616          & 0.751          & 0.482          &0.587          & 0.685          \\ \hline
OverlapTransformer~\cite{ma2022ral}        & 0.861          & 0.899          & 0.930          & 0.639          & 0.697          & 0.780          & 0.536          & 0.645          & 0.764          & 0.496          & 0.603          & 0.715          \\ \hline
FusionVLAD~\cite{yin2021fusionvlad}        & 0.786          & 0.870          & 0.922    & 0.462          & 0.579      & 0.702    & 0.510          & 0.643          & 0.754  & 0.429          & 0.553          & 0.667                   \\ \hline
CVTNet (ours)              & \textbf{0.932} & \textbf{0.946} & \textbf{0.957} & \textbf{0.793} & \textbf{0.835} & \textbf{0.871} & \textbf{0.817} & \textbf{0.866}          & \textbf{0.899} & \textbf{0.780} & \textbf{0.826} & \textbf{0.869} \\
\bottomrule
\end{tabular}
  \label{tab:pr_on_nclt}
  \vspace{-0.2cm}
\end{table*}


\begin{table*}[t]
\center
\setlength{\tabcolsep}{10.3pt}
\renewcommand\arraystretch{1.2}
\caption{Comparison of recognition performance on the KITTI and self-recorded datasets}
\begin{tabular}{lcccccccc}
\toprule
\multirow{2}{*}{Approach} & \multicolumn{3}{c}{Sequence-2}               & \multicolumn{3}{c}{Sequence-3 (reverse)}                     & \multicolumn{2}{c}{KITTI}                     \\ \cline{2-9} 
                          & AR@1        & AR@5        & AR@20 & Recall@1        & AR@5        & AR@20       & AUC        & F1max               \\ \hline
Scan Context~\cite{kim2018scan}              & 0.877          & 0.958          & 0.986          & 0.564          & 0.710          & 0.810          & 0.836          & 0.835         \\ \hline
PointNetVLAD~\cite{uy2018pointnetvlad}              & 0.913          & 0.936          & 0.957    & 0.400          & 0.521          & 0.623          & 0.856          & 0.846                    \\ \hline
MinkLoc3D~\cite{Komorowski2021wacv}        & 0.948          & 0.971          & 0.993    & 0.450          & 0.653          & 0.808          & 0.894          & 0.869                   \\ \hline
OverlapTransformer~\cite{ma2022ral}        & 0.974          & 0.987          & 0.995    & 0.776          & 0.878          & 0.931          & 0.907          & 0.877                   \\ \hline
FusionVLAD~\cite{yin2021fusionvlad}        & 0.971          & 0.990          & 0.997    & 0.865          & 0.923          & 0.960          & 0.882          & 0.860                   \\ \hline
CVTNet (ours)              & \textbf{0.992} & \textbf{0.996} & \textbf{0.998}    & \textbf{0.907} & \textbf{0.951} & \textbf{0.993} & \textbf{0.911} & \textbf{0.880}   \\
\bottomrule
\end{tabular}
  \label{tab:pr_on_haomo}
\end{table*}


The first experiment supports our claim that our approach achieves state-of-the-art place recognition and loop closure detection in outdoor large-scale environments using multi-view LiDAR data.

Following~\cite{ma2022ral}, we use the NCLT dataset for evaluating long-term place recognition, where sequence 2012-01-08 is used to train the models and sequences 2012-02-05, 2012-06-15, 2013-02-23, and 2013-04-05 are used to evaluate the recognition performance. \change{Besides NCLT dataset, we also recorded our own dataset with multiple reverse driving conditions repeated in the same environment, which are frequent cases in autonomous driving but challenging to place recognition methods.} We use sequence-1 as the database for training and two other runs (sequence-2 and sequence-3) as query sequences for evaluation. They all share the same route but driving from different directions. In contrast, KITTI odometry sequences are used for evaluating loop closure detection, where we use sequences~$03$--$10$ for training, sequence~$02$ for validation, and sequence~$00$ for evaluation.

We use average top 1 recall (AR@1), top 5 recall (AR@5), and top 20 recall (AR@20) as the metrics of the evaluation conducted on NCLT and the self-recorded datasets. We also use AUC and F1 max scores as the evaluation metrics for the KITTI dataset as introduced in~\cite{ma2022ral}. The evaluation results are shown in \tabref{tab:pr_on_nclt} and \tabref{tab:pr_on_haomo}
As can be seen, our proposed CVTNet has better performance than all the state-of-the-art baseline methods on the place recognition metrics. Compared to OverlapTransformer, CVTNet utilizes the combination of RIVs and BEVs as input rather than only range images, which increases the performance of recognizing places with large time gaps by $7.1\% \sim 28.4\%$ on AR@1 of the NCLT dataset. In contrast to FusionVLAD which also considers different views, our proposed CVTNet outperforms it by $14.6\% \sim 35.1\%$ on AR@1 of the NCLT dataset, and $2.1\% \sim 4.2\%$ on AR@1 of the self-recorded dataset. For loop closure detection on the KITTI dataset, our method also significantly outperforms all baseline methods.

\subsection{Ablation Study on Transformer Modules}
\label{sec:ablation_on_trans}

\begin{figure}[t]
  \centering
  \includegraphics[width=0.9\linewidth]{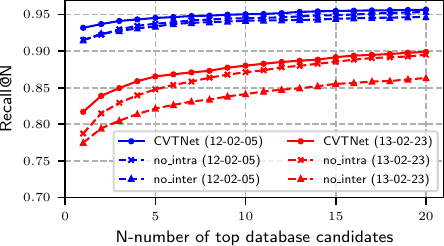}
  \caption{Ablation study on intra- and inter-transformer mudules.}
  \label{fig:aba_on_inter_intra_on_nclt}
  \vspace{-0.4cm}
\end{figure}

This ablation study investigates the effect of intra- and inter-transformer modules. We compare our CVTNet with the variants of not using transformer modules.
We evaluate all variants using the average recall AR@$N$ on sequence 2012-02-05 and 2013-02-23 of the NCLT dataset. The results are shown in \figref{fig:aba_on_inter_intra_on_nclt}. As can be seen, our CVTNet with both intra- and inter-transformer modules consistently outperforms all variants. Besides, CVTNet with only an inter-transformer outperforms the one using only an intra-transformer, which indicates that the inter-transformer contributes more to enhancing the discrimination of the final global descriptor than the intra-transformer. The benefit of using two types of transformers is more prominent in place recognition with a larger time gap, which can be drawn by comparing the results between two sequences recorded from different times.

\vspace{-0.4cm}
\change{\subsection{Ablation Study on Cross-Transformer Blocks}}
\label{sec:ablation_on_ctb}

\begin{table*}[h]
\setlength{\tabcolsep}{3.6pt}
\center
\renewcommand\arraystretch{1.2}
\caption{Comparison of CVTNet with different numbers of cross-transformer blocks on the NCLT dataset}
\begin{tabular}{cccccccccccccc}
\toprule
\multirow{2}{*}{Number} & \multirow{2}{*}{Runtime [ms]}  & \multicolumn{3}{c}{2012-02-05}                      & \multicolumn{3}{c}{2012-06-15}                      & \multicolumn{3}{c}{2013-02-23}                      & \multicolumn{3}{c}{2013-04-05}                      \\ \cline{3-14} 
                    &        & AR@1        & AR@5        & AR@20       & AR@1        & AR@5        & AR@20       & AR@1        & AR@5        & AR@20       & AR@1        & AR@5        & AR@20       \\ \hline
$1$    &5.46    & 0.924          & 0.943          & 0.955          & 0.781          & 0.826          & 0.860          & 0.798          & 0.843          & 0.893          & 0.749          & 0.797          & 0.840          \\ \hline
$2$   &6.49          & \textbf{0.932} & \textbf{0.946} & \textbf{0.957} & \textbf{0.793} & \textbf{0.835} & \textbf{0.871} & \textbf{0.817} & \textbf{0.866}          & \textbf{0.899} & \textbf{0.780} & \textbf{0.826} & \textbf{0.869} \\
\hline
$3$     &7.32         & 0.923 & 0.940 & 0.955 & 0.783 & 0.824 & 0.861 & 0.787 & 0.840  & 0.889 & 0.757 & 0.807 & 0.856 \\
\bottomrule
\end{tabular}
  \label{tab:ablation_trans}
  \vspace{-0.2cm}
\end{table*}

This ablation study aims to show the effectiveness of the number of cross-transformer blocks in the inter-transformer module of CVTNet. Note that we only compare our method with $1\sim3$ cross-transformer blocks to maintain low memory consumption.
The experimental results on the NCLT dataset are shown in~\tabref{tab:ablation_trans}. As can be seen, the inter-transformer with two cross-transformer blocks outperforms the ones with one and three cross-transformer blocks. The reason could be that
more transformer blocks might need more training data and time to obtain good performance. We also show the average runtime to generate one global descriptor for sequence 2012-01-08 of the NCLT dataset in~\tabref{tab:ablation_trans}. It can be seen that more cross-transformer blocks cost more time to extract features. Thus, we use $2$ transformer blocks in our CVTNet to keep effectiveness and efficiency. 


\vspace{0.5cm}

\subsection{Study on Multi-Layer Input}
\label{sec:study_on_multi_layer}

In this experiment, we use self-recorded datasets to validate the superiority of using multi-layer input for CVTNet. We modify the input channel number of the first layer of OverlapNetLeg to receive the single-channel input.
The experimental results are shown in \tabref{tab:multi}. As can be seen, CVTNet with multi-layer input outperforms the baseline with only single-channel input. Our proposed multi-layer input explicitly tackles the problem of occluded laser points in the background, providing more input point cloud information, thus enabling the network to reweight data from different subspaces to enhance the discrimination of the extracted features.

To further validate the effectiveness of the proposed multi-layer form of inputs, we also apply it to another network, FusionVLAD as also shown in \tabref{tab:multi}. The multi-layer input form also improves the performance of FusionVLAD, which indicates that our proposed multi-layer input form generalizes well into different methods.
Still, our method significantly outperforms the FusionVLAD with multi-layer inputs due to the devised novel intra- and inter-transformer network.
\begin{table}[t]
  \centering
  \setlength{\tabcolsep}{10pt}
  \renewcommand\arraystretch{1.2}
  \caption{Study on multi-layer input}
\begin{tabular}{L{100pt}|C{20pt}C{20pt}C{20pt}}
\toprule
Approach         & AR@1               & AR@5            & AR@20                 \\ \hline
FusionVLAD (single-channel)             & 0.865               & 0.923            & 0.960        \\ \hline
FusionVLAD (multi-layer)                    & 0.868            & 0.930             & 0.970           \\  \hline
CVTNet (single-channel)             & 0.896               & 0.948            & 0.987                  \\ \hline
CVTNet (multi-layer)                    & \textbf{0.907}             & \textbf{0.951}             & \textbf{0.993}                      \\ 
\bottomrule
\end{tabular}
  \label{tab:multi}
  \vspace{-0.4cm}
\end{table}

\subsection{Study on Yaw-Angle Invariance}
\label{sec:yaw-angle-invariance}

\begin{figure}[t]
  \vspace{0.4cm}
  \centering
  \includegraphics[width=0.9\linewidth]{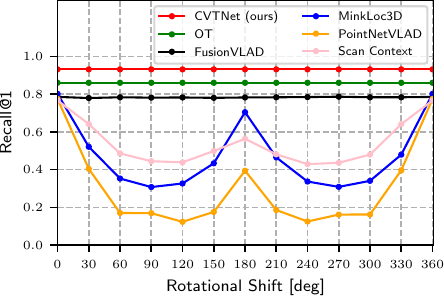}
  \caption{\change{Study on yaw-angle invariance.}}
  \label{fig:yaw_angle_inv_fig}
    \vspace{-0.4cm}
\end{figure}

We have shown the robustness of our proposed CVTNet to viewpoint changes using the self-recorded dataset in \secref{sec:epr}. In this experiment, we further design a more straightforward way to validate that our devised network architecture is strictly invariant to the yaw rotation of LiDAR scans obtained at the same place. We rotate each query scan of sequence 2012-02-05 along the yaw-axis in steps of 30 degrees and search the same places in the database built from sequence 2012-01-08. We compare our method with the state-of-the-art baseline methods and use Recall@1 as the criterion to show the effect of yaw angle changes on place recognition performance. The experimental results are shown in \figref{fig:yaw_angle_inv_fig}. 
As can be seen, CVTNet outperforms the other methods consistently in terms of Recall@1 in all the rotation conditions.
Both CVTNet and OverlapTransformer are not affected by the rotation along the yaw-axis due to the devised yaw-rotation-invariant architectures. FusionVLAD is not significantly affected by yaw-angle rotation but is not strict yaw-rotation-invariant. In contrast, the performance of MinkLoc3D and PointNetVLAD decreases fast with the yaw angle increasing, which means their network architectures are not robust enough to viewpoint changes.

\subsection{Runtime and Memory Consumption}
\label{sec:runtime}

\begin{figure}[t]
  \vspace{0.4cm}
  \centering
  \includegraphics[width=0.9\linewidth]{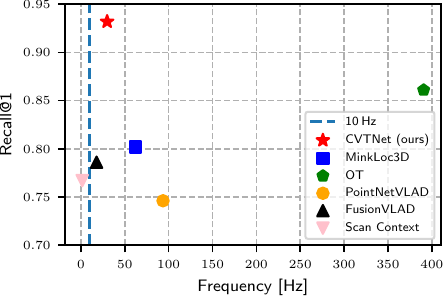}
  \caption{\change{Comparison of runtime with state-of-the-art methods.}}
  \label{fig:runtime_recall}
    \vspace{-0.4cm}
\end{figure}

\change{We further compare the real-time performance of our CVTNet with all the baseline methods, and evaluate its memory consumption. We conduct the experiments on a system with an Intel i7-11700K CPU and an Nvidia RTX 3070 GPU. We use NCLT sequence 2012-01-08 as database and sequence 2012-02-05 as query to calculate the total runtime of global descriptor generation and top $20$ candidates retrieval within $28127$ database laser scans for each query laser scan. FAISS library~\cite{johnson2019billion} is used here to accelerate the searching process except for Scan Context which can only use an ad-hoc retrieval function for shift-based matching~\cite{kim2018scan}. \figref{fig:runtime_recall} shows the running frequency v.s. top 1 recall. The running frequency of CVTNet is $29.85$\,Hz ($33.50$\,ms in total, $15.03$\,ms for descriptor generation, and $18.47$\,ms for retrieval) which is about 3 times faster than a typical LiDAR sensor frame rate at $10$\,Hz, while achieving the state-of-the-art place recognition performance. In addition, the number of parameters of CVTNet is $21.38$\,M, showing the small memory consumption of our approach.}


\section{Conclusion}
\label{sec:conclusion}
%
\changesec{In this article, a novel cross-view transformer network has been presented for LiDAR-based place recognition. Point clouds have been projected to multi-layer RIVs and BEVs as the inputs of our method instead of the mundane single-channel form. The intra-transformer has been utilized in the proposed network to capture the inner correlation within the individual view, while the inter-transformer has been used to extract the cross-view correlation between different views. The experimental comparisons of our method with the state-of-the-art methods have validated its superior performance in both place recognition and loop closure detection. Moreover, the experimental results have also shown that our proposed CVTNet is yaw-angle-invariant against viewpoint changes and operates online which can be used for real autonomous driving applications.}


\bibliographystyle{IEEEtran}

\footnotesize{
\bibliography{new}}

\end{document}


\maketitle

\subsection{Pipeline of generating global descriptors using CVTNet}
We provide an additional pseudocode diagram showing how to generate a 1-D global descriptor for one input 3-D point cloud using our proposed approach in \algref{alg:pipline}. The multi-view multi-layer generator (MMG) first splits the region by range and height intervals, and conducts spherical and top-down projection respectively to generate multi-layer input data. The aligned feature encoder (AFE) combines the multi-layer RIVs and BEVs as inputs and uses OverlapNetLeg to generate sentence-like features. It then applies an intra-transformer to extract the inner correlation of the compressed features of each view. The multi-view fusion module (MVF) uses an inter-transformer to fuse the aligned features from different views and then applies NetVLAD with multi-layer perceptrons (MLPs) on both single-view and fused features to generate the final 1-D global descriptor.

\begin{algorithm}[h]
\small
\caption{The pipeline of generating the descriptor using CVTNet}
\hspace*{0.02in} {\bf Input:}
3-D point cloud $\mathcal{P}$, range intervals $\{s_0, s_1, s_2, s_3, ..., s_q\}$, height intervals $\{t_0, t_1, t_2, t_3, ..., t_q\}$;\\
\hspace*{0.02in} {\bf Output:}
1-D global descriptor $\mathcal{G}$;
\begin{algorithmic}[1]
\State MMG discretizes the region using the range intervals to obtain the split space $E^r=\{E_1^r, E_2^r, E_3^r, ..., E_q^r\}$;
\For{each subspace $E_i^r \in E^r$}
   \State Collect the point cloud $\mathcal{P}_i$ in $E_i^r$;
   \State Apply spherical projection on each laser points $\vec{p}_{k} \in \mathcal{P}_i$ with Eq.~(1);
   \State Generate the range image view $\mathcal{R}_i$ for $\mathcal{P}_i$ based on the results of spherical projection;
\EndFor
\State MMG discretizes the region using the height intervals to obtain the split space $E^b=\{E_1^b, E_2^b, E_3^b, ..., E_q^b\}$;
\For{each subspace $E_j^b \in E^b$}
   \State Collect the point cloud $\mathcal{P}_j$ in $E_j^b$;
   \State Apply top-down projection on each laser points $\vec{p}_{k} \in \mathcal{P}_j$ with Eq.~(2);
   \State Generate the bird's eye view $\mathcal{B}_j$ for $\mathcal{P}_j$ based on the results of top-down projection; 
\EndFor
\State Concatenate all the generated range image views and bird's eye views along the channel dimension respectively and get $\mathbb{R}=\{\mathcal{R}_i\}$ and $\mathbb{B}=\{\mathcal{B}_j\}$;
\State AFE extracts the aligned features $\mathcal{A}_0^r$ for $\mathbb{R}$ and $\mathcal{A}_0^b$ for $\mathbb{B}$ using two branches with OverlapNetLeg and intra-transformers;
\State MVF fuses $\mathcal{A}_0^r$ and $\mathcal{A}_0^b$ based on the inter-transformer and generates $\mathcal{A}_l^f$;
\State The combination of NetVLAD and MLP converts $\mathcal{A}_0^r$, $\mathcal{A}_0^b$, and $\mathcal{A}_l^f$ to three 2D vectors $g^r, g^b, g^{f}$ each with size of $1\times 256$;
\State Concatenate $[g^r, g^b, g^{f}]$ to generate the final global descriptor $\mathcal{G}$ with size of $1\times 768$;
\State \Return $\mathcal{G}$.
\end{algorithmic}
\label{alg:pipline}
\end{algorithm}

\subsection{Complexity Analysis}
We present the results of the complexity analysis in \tabref{tab:complexity_analysis}. We use GFLOPs to analyze the time complexity and use the number of network parameters to analyze the space complexity. Because MinkLoc3D utilizes MinkowskiEngine proposed by NVIDIA, its time complexity cannot be calculated using any existing tool. Thus, we do not provide GFLOPs of MinkLoc3D in the experiment. 
It can be seen that the number of parameters of our proposed CVTNet is less than the other multi-view LPR method FusionVLAD, and is slightly greater than PointNetVLAD and OverlapTransformer. Although our proposed CVTNet has larger GFLOPs than the baseline methods, it achieves the state-of-the-art place recognition performance with around $30$\,Hz in the runtime experiment in Sec.~IV-G. 
\begin{table}[h]
\centering
\setlength{\tabcolsep}{15pt}
\renewcommand\arraystretch{1.4}
\caption{Complexity analysis}
\begin{tabular}{lcc}
\toprule
Approach         & GFLOPs    & Params \\ \hline
PointNetVLAD~\cite{uy2018pointnetvlad}                   & 3.58            & 19.78\,M    \\  \hline
MinkLoc3D~\cite{Komorowski2021wacv}                       & -            & 5.28\,M     \\  \hline
OverlapTransformer~\cite{ma2022ral}                       & 2.61            & 18.43\,M \\  \hline
FusionVLAD~\cite{yin2021fusionvlad}                      & 6.15             & 54.20\,M  \\  \hline 
CVTNet (ours)                       & 8.23             & 21.38\,M     \\ 
\bottomrule
\end{tabular}
\label{tab:complexity_analysis}
\end{table}

\subsection{SLAM Results}
We present in \figref{fig:loam_liosam} the SLAM results with and w/o our proposed CVTNet closing loops, and the ground truth global map. We integrate CVTNet into LIO-SAM~\cite{shan2020lio} and visualize the mapping results on KITTI sequence 08 (raw data 2011\_09\_30\_0028) \cite{geiger2012cvpr}. \figref{fig:loam_liosam} shows that CVTNet helps LIO-SAM to close loops and generate more global-consistent map compared to the ground truth map, which demonstrates the feasibility of using our proposed method in SLAM systems. Note that the ground truth map is built by merging all the point clouds of KITTI sequence 08 with provided ground truth poses.
\begin{figure}[t]
  \centering
  \subfigure[LIO-SAM with CVTNet]{\includegraphics[width=0.8\linewidth]{pics_seqot/crop_kitti08_with_loop.png}} 
  \subfigure[LIO-SAM without CVTNet]{\includegraphics[width=0.8\linewidth]{pics_seqot/crop_kitti08_without_loop.png}}
    \subfigure[Ground Truth]{\includegraphics[width=0.8\linewidth]{pics_seqot/crop_kitti08_gt.png}}  
  \caption{Comparison of SLAM results with and w/o using CVTNet to close loops}
  \label{fig:loam_liosam}
\end{figure}

\bibliographystyle{IEEEtran}
\footnotesize{
\bibliography{glorified,new}}